# Ask1: Development and Reinforcement Learning-Based Control of a Custom Quadruped Robot

Yang Zhang, Yuxing Lu, Guiyang Xin*, Yufei Xue, Chenkun Qi, Kairong Qin, Yan Zhuang

***Abstract*—In this work, we present the design, development, and experimental validation of a custom-built quadruped robot, Ask1. The Ask1 robot shares similar morphology with the Unitree Go1, but features custom hardware components and a different control architecture. We transfer and extend previous reinforcement learning (RL)-based control methods to the Ask1 robot, demonstrating the applicability of our approach in real-world scenarios. By eliminating the need for Adversarial Motion Priors (AMP) and reference trajectories, we introduce a novel reward function to guide the robot's motion style. We demonstrate the generalization capability of the proposed RL algorithm by training it on both the Go1 and Ask1 robots. Simulation and real-world experiments validate the effectiveness of this method, showing that Ask1, like the Go1, is capable of navigating various rugged terrains.**

## I. INTRODUCTION

Quadruped robots offer significant potential for a wide range of applications due to their ability to navigate diverse terrains with stability and agility. With advancements in multi-legged control systems and bio-inspired design, these robots are becoming increasingly suited for tasks such as disaster response, environmental monitoring, and automated transport. Their adaptability in complex and unpredictable environments makes them a valuable asset for both practical applications and the advancement of intelligent systems.

In traditional quadruped robot control, one of the most common techniques is Model Predictive Control (MPC) [1], [2], [3], which ensures stability by optimizing the robot's motion in real-time based on simplified dynamic models and predefined contact sequences. MPC performs well on flat terrains, but its adaptability and computational efficiency are limited when faced with more complex environments, such as uneven terrains or obstacles. To address this challenge,

Yang Zhang, Guiyang Xin, Kairong Qin are with the School of Biomedical Engineering, Dalian University of Technology, Dalian 116024, China (e-mail: guiyang.xin@dlut.edu.cn).

Yuxing Lu is with the School of Computer Science and Technology, Dalian University of Technology, Dalian 116024, China (e-mail: 2021102042@mail.dlut.edu.cn)

Yufei Xue, Chenkun Qi are with the School of Mechanical Engineering, Shanghai Jiao Tong University, Shanghai 200240, China (e-mail: xue_yeiii@sjtu.edu.cn; chenkqi@sjtu.edu.cn).

Yan Zhuang is with the School of Control Science and Engineering, Dalian University of Technology, Dalian 116024, China (e-mail: zhuang@dlut.edu.cn).

*Corresponding author

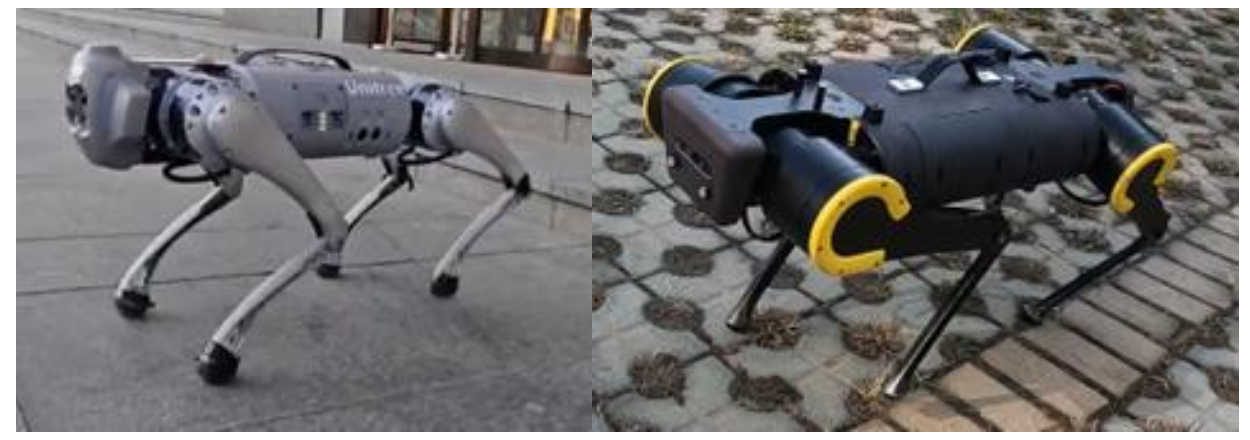

**Fig. 1.** Unitree Go1 (left) and Ask1(right). Video is available at https://youtu.be/_KOv6pzVHx8.

many methods decompose the problem into two sub-tasks: contact planning and motion optimization [4], [5]. However, this still requires significant expertise and does not guarantee optimal performance in real-world scenarios.

In contrast, RL [10]-[16] methods have gained widespread application in recent years, particularly in robotics. RL allows
robots to learn optimal control policies through simulation training, based on interactions with the environment, without relying on pre-designed models or predefined contact sequences. This provides greater flexibility and adaptability, especially in dynamic and complex environments. For example, RL can learn directly from sensor data, avoiding the need for prior modeling of environmental parameters, and enables online adjustment of control strategies to quickly adapt to unseen terrains or task variations. This approach not only improves performance on flat ground but also helps the robot cope with real-world complexities such as stairs, rough terrains, and external disturbances. Through this method, RL offers a more general and adaptive control framework for quadruped robots, overcoming the limitations of traditional control techniques in handling diverse and unpredictable environments.

In our previous work [7], the teacher-student framework involved training the teacher policy first, followed by the student policy. In this setup, the teacher network had access to privileged information and used AMP [9] to enforce a specific motion style, while the student policy learned through imitation learning from the teacher's behavior. In another study [6], we employed an asymmetric framework [18] for training, where the actor network only used sensory information available to the robot, while the critic network had access to privileged information. AMP was again used to constrain the motion style. Both of these approaches were validated on the Unitree Go1 quadruped robot.

In this work, we extended our previous training methods to our custom-built quadruped robot, Ask1. We introduced a new training framework that eliminates the need for AMP and the collection of reference trajectories. Instead, we use a novel reward function to achieve similar motion style constraints. We validated this approach in both simulation and real-world environments, demonstrating promising results on both the Unitree Go1 and Ask1 robots. Images of the Ask1 and Unitree Go1 robots are shown in Fig. 1.

TABLE I
MECHANICAL PARAMETERS

| Name | Payload [kg] | Mass [kg] | Thigh length [m] | Calf length [m] | Size [mm] | Power supply [VDC] |
|---|---|---|---|---|---|---|
| Unitree Go1 | 5 | 12 | 0.23 | 0.24 | 645*280*400 | 24 |
| Ask1 | 10 | 20 | 0.28 | 0.25 | 840*360*430 | 48 |

The remaining of the paper is organized as follows. In Section II, the hardware design of the Ask1 is introduced. The proposed RL method is described in Section III. In Section IV, several experiments are conducted to validate the effectiveness of the proposed approach. We conclude the paper in Section V.

## II. DESIGN OF THE ASK1 QUADRUPED ROBOT

We designed and developed a new quadruped robot, Ask1, as shown in Fig. 2. Ask1 shares the same morphology as Go1 but features a larger overall size, as detailed in Table I. Each leg of Ask1 has three degrees of freedom, with the knee joint actuated by a four-bar linkage mechanism. Unlike some quadruped robots, Ask1 does not incorporate contact force sensors on its feet. The robot's 12 joints are powered by low-backlash actuators integrated with planetary reducers, each capable of delivering a maximum torque of 25 Nm.

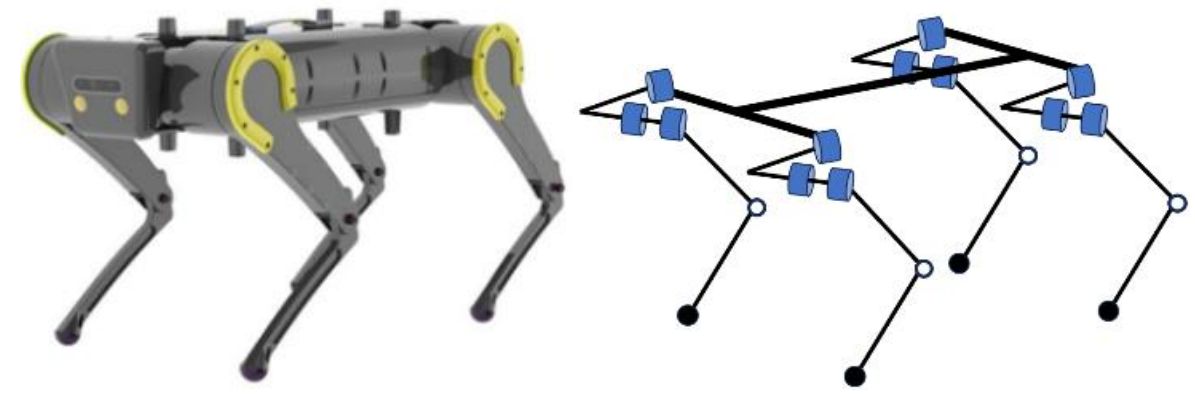

**Fig. 2.** 3D rendering (left) and kinematic tree (right) of Ask1.

Figure 3 illustrates the system diagram of the Ask1 robot. The onboard control PC is an Intel NUC, which runs the control policy trained by RL algorithms to track velocity commands from the joystick. Joint data is transmitted via USB to CAN bus adapters, facilitating communication between the control PC and the DC motors. The 12 joint actuators are highly integrated drive units, combining brushless DC motors, planetary reducers, encoders, and drive boards into a single unit. The IMU on the Ask1 is an Xsens MTi-630, which measures the angular velocities and linear accelerations of the torso and provides this data to the control policy. A dedicated NVIDIA Jetson Orin NX processes images and point clouds from an Intel Realsense camera and a Robosense LiDAR to assist in robot navigation. The two onboard PCs communicate through an onboard router. It is important to note that the camera and LiDAR were not used in the experiments presented in this paper, as the robot operated without visual input.

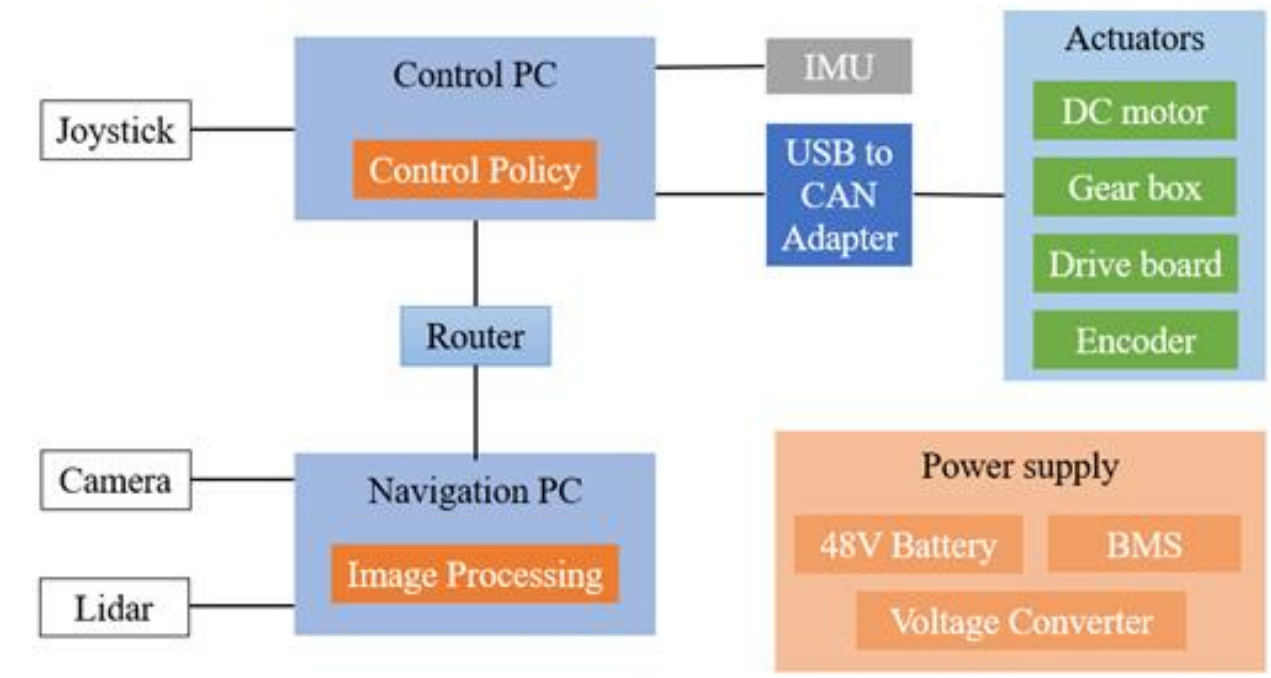

**Fig. 3.** System diagram of Ask1.

## III. METHOD

### A. *Reinforcement Learning Problem Formulation*

RL is a key branch of machine learning where an agent learns to make decisions by interacting with the environment, with the goal of maximizing long-term rewards. The core idea of RL is that the agent learns an optimal policy through trial and error, adjusting its actions based on the feedback (rewards or penalties) received from the environment. In RL, the environment is represented by a state space, and the agent chooses actions based on the current state, transitioning to the next state and receiving a reward. The goal of RL is to find an optimal policy that maximizes the expected cumulative reward over time:

$$J(\theta)=E_{\pi_\theta}\left[\sum_{t=0}^{\infty}\gamma^t r_t\right] \quad , \tag{1}$$

where $\gamma^t$ is a discount factor.

In environments with partially observable states, such as robot control problems that rely on sensor inputs, the problem can be modeled as a Partially Observable Markov Decision Process (POMDP) [7], [10], [11], [12]. In this case, the agent cannot fully observe the environment, relying instead on available sensory data and historical information to make decisions. To tackle this challenge, we employ the Proximal Policy Optimization (PPO) [14] algorithm within the context of a partially observable environment. Specifically, we utilize the asymmetric actor-critic framework, consistent with our previous work [6], where the agent's decision-making process is structured to improve learning efficiency and stability, even in the face of limited

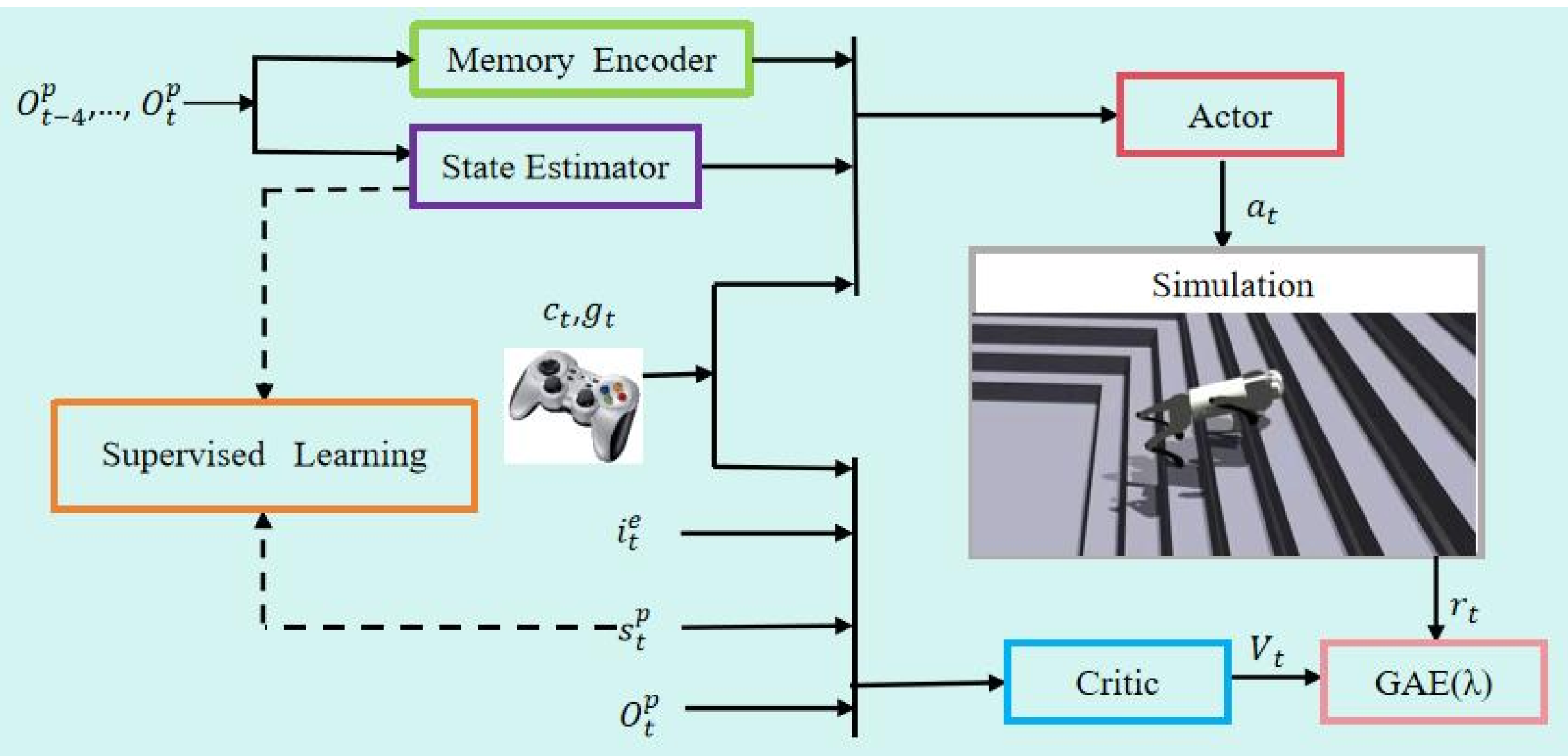

**Fig. 4.** Overview of the learning framework. The actor generates actions based on partial observations, while the critic evaluates the policy using the full state, including privileged information. This structure allows for real-world deployment with enhanced performance through simulation-based data.

or incomplete state observations.

*State Space*: In this problem, the critic network's state space includes a full state consisting of partial observations ( $o^p$ ) includes gravity projection components, base angular velocity, joint positions, joint velocities. User commands include two components: $c_t$ and $g_t$ . $c_t$ represents the desired velocity commands, including the longitudinal, lateral, and yaw velocities, while $g_t$ contains the remaining gait information, such as phase information for all four legs, step frequency, body height, and the standing/swing ratio. These components are used to control the robot's motion. Privileged state ( $s^p$ ) includes the base linear velocity, contact forces, ground friction coefficients, and other factors that are typically only available in simulation, which help the agent better estimate the current state. Terrain information ( $i^e$ ) consists of a height map of the robot's surroundings, representing the terrain's vertical variation at 187 sampled points around the robot. This asymmetric structure is advantageous because it allows the policy to be deployed on a real robot using only onboard sensors (with $o^p$ and command) while still exploiting the privileged information ( $s^p$ and $i^e$ ) available in simulation to improve training. This approach ensures better performance when the policy is transferred from simulation to real-world environments.

*Action Space*: The action space is represented as a 12-dimensional vector, where each element corresponds to a joint position offset. These offsets are added to the nominal joint positions when the robot is in a standing position, forming the target joint positions. Specifically, the target joint positions are calculated as follows: The 12-dimensional vector outputs the joint position offsets, which, when added

( $o^p$ ),user commands, privileged state information ( $s^p$ ), and terrain information ( $i^e$ ). Specifically, Partial observation to the nominal joint positions, form the target motor positions. These target joint positions are then passed to low-level controllers, which compute the torque commands using PD controllers with fixed gains (Kp = 20, Kd = 0.5).

### *B. Network Architecture*

The overall architecture of our learning framework is shown in Fig. 4. Both the actor and critic networks use a multi-layer perceptron (MLP) structure, with Exponential Linear Unit (ELU) as the activation function for the hidden layers. Specifically, the actor network consists of a state estimator, a memory encoder, and a low-level MLP. These components work together to generate actions from partial observations and commands. The critic network evaluates the current policy by estimating the value of the current state, providing feedback for policy updates. More details on each layer are shown in Table II.

To handle the history of partial observations, we use an MLP to encode the sequential correlations between short-term histories. The network receives the five most recent partial observations $o_{i-4}^P,\ldots,o_{i-1}^P,o_i^P$ as input and outputs an encoded history $h_t \in \mathbb{R}^{32}$ , which helps the robot implicitly estimate the privileged state and terrain information. Along with this, the state estimator concurrently estimates the base linear velocity $\hat{v}_t$ using supervised learning. The $\hat{v}_t$ helps the low-level network understand how well the locomotion task is being performed under the current command $c_t$ , while $h_t$ provides awareness of the interactions between the robot and the environment.

The $h_t$ , $\hat{v}_t$ , $c_t$ ,and $g_t$ are then concatenated and fed into the low-level network with a tanh output layer to generate the mean $\mu_t \in \mathbb{R}^{32}$ of a Gaussian distribution $a_t \sim N(\mu_t, \sigma)$ ,where $\sigma \in \mathbb{R}^{12}$ denotes the variance of the action determined by PPO. This network architecture is effective in learning from limited observations and continuously improving the policy through interactions with the environment, ultimately enhancing the robot's performance in complex scenarios.

TABLE II
NETWORK ARCHITECTURES

| Module | Inputs | Hidden Layers | Outputs |
|---|---|---|---|
| History Encoder | $o_{i-4}^{P},...,o_{i-1}^{P},o_{i}^{P}$ | [256,128,64] | $h_t$ |
| State Estimator | $o_{i-4}^{P},...,o_{i-1}^{P},o_{i}^{P}$ | [256,128,64] | $v_t$ |
| Actor | $h_t, v_t, c_t, g_t$ | [256,128,64] | $a_t$ |
| Critic | $c_t, g_t, i_t^{e}, s_t^{p}, o_t^{p}$ | [512,256,128] | $V_t$ |

### *C. Reward Functions Design*

The reward function consists of task rewards and style rewards. The task reward $r^g$ includes velocity tracking in three directions, aiming to ensure the robot's control over linear and angular velocities during task execution. The style reward is divided into three components: $r^l, r^s$ , and $r^c$ , which collectively guide the robot's motion style optimization. We adopt the same approach as in previous work [6] to design $r^c$ in order to achieve the desired gait.

Inspired by the Raibert Heuristic [8], we enhance the calculation of foot positions by incorporating adjustments to the stance width and length to align with the desired contact schedule and body velocity. More details can be found in Fig.5. Specifically, the foot positions in the ground plane are computed by adjusting the nominal stance based on phase-dependent offsets. Let $p_{f,x,y}^{\text{desired}}$ represent the desired foot positions in the x- and y-coordinates for each foot. These positions are defined as:

$$p_{f,x,y}^{\text{desired}} = p_{f,x,y}^{\text{norm}} + \triangle p_{f,x,y} , \tag{2}$$

where $p_{f,x,y}^{\text{norm}}$ are the nominal foot positions for each foot (e.g., [FR, FL, RR, RL]) based on the stance width and length, and $\triangle p_{f,x,y}$ are the phase-dependent offsets.

The offset for the x- and y-coordinates is determined by the desired velocities of the robot, including both linear and angular velocities, and the gait frequency. For the y-direction (lateral), the offset is determined by the combination of the commanded lateral velocity $v_v^{\text{cmd}}$ and the lateral velocity $v_v^{\text{yaw}}$ induced by the robot's rotational yaw velocity $\omega_z$ . These two components are both modulated by the gait frequency $f$ and the phase $\phi$ of each foot. The phase of each foot oscillates between stance and swing phases, represented as $\phi$ .The lateral offset for each foot is

$$\begin{aligned} \triangle p_{f,y} &= \phi \frac{v_y^{\text{cmd}}}{f} + \phi \frac{v_y^{\text{yaw}}}{f} , \\ v_x^{yaw} &= \frac{\omega_z W}{2} , \end{aligned} \tag{3}$$

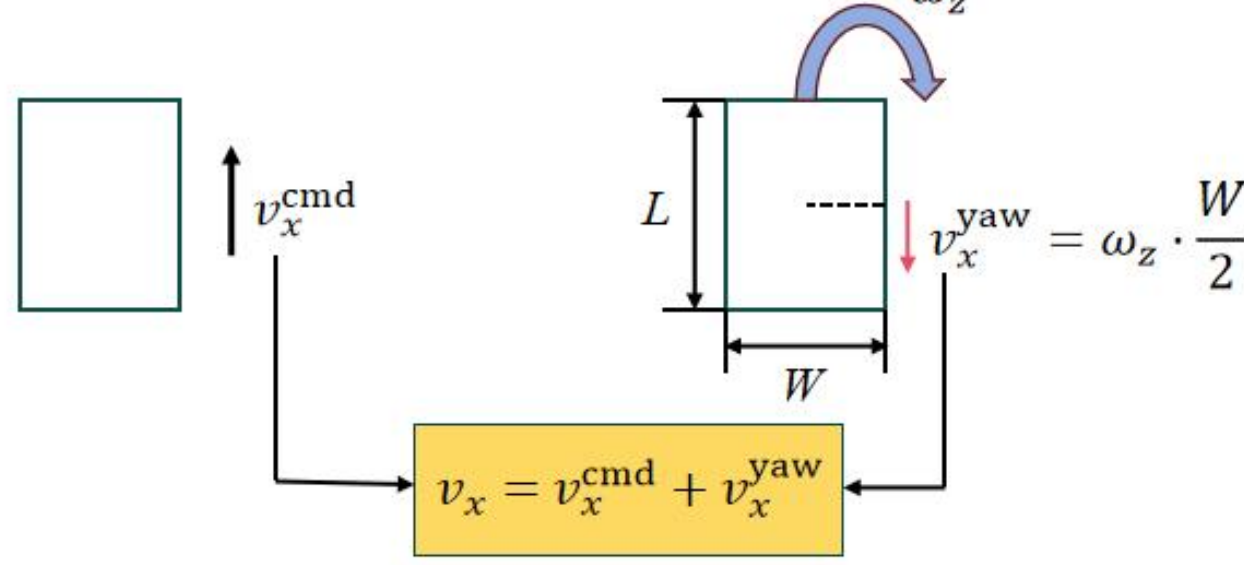

**Fig. 5.** Illustration of Raibert's calculation method for determining foot positions, incorporating adjustments based on stance width, length, and phase-dependent offsets to optimize gait stability.

where $v_v^{\text{cmd}}$ is the desired lateral velocity , $\omega_z$ is the desired angular velocity around the vertical axis (yaw), $L$ is the desired stance length, $f$ is the gait frequency. Similarly, for the x-direction (forward), the offset is:

$$\begin{aligned} \triangle p_{f,x} &= \phi \frac{v_x^{\text{cmd}}}{f} + \phi \frac{v_x^{\text{yaw}}}{f} , \\ v_x^{\text{yaw}} &= \frac{\omega_z W}{2} , \end{aligned} \tag{3}$$

where $v_x^{\text{cmd}}$ is the commanded forward velocity, $\omega_z$ is the desired angular velocity around the vertical axis (yaw), $W$ is the desired stance width, $f$ is the gait frequency.

The reward is then calculated by comparing the actual foot positions $p_{f,x,y}^{\text{actual}}$ to the desired foot positions. The error is computed as the sum of squared differences:

$$r_s = \sum_i \sum_j (p_{f,x,y}^{\text{desired}} - p_{f,x,y}^{\text{actual}}) , \tag{5}$$

where $i$ and $j$ iterate over the four feet and their respective x and y coordinates. This reward function helps the robot minimize the gap between actual and desired foot positions, leading to a stable and smooth gait. By using this reward, the robot can achieve natural movements even without AMP. The details of the reward functions are shown in Table III. The total reward $r_t$ is defined by:

$$r_t = r_t^g + r_t^l + r_t^s + r_t^c , \tag{6}$$

## IV. EXPERIMENTS

### *A. Training Setup*

We trained 4096 parallel agents on different types of terrains using the IsaacGym simulator [13] and a single NVIDIA RTX 4090 GPU. Each RL episode has a maximum duration of 20 seconds, and it terminates earlier if specific termination conditions are met, such as when the trunk makes contact with the ground or when the base orientation deviates from the normal range. The control frequency of the policy is set to 50 Hz during simulation. To enhance the generalization of the policy and facilitate its transfer from simulation to real-world applications, we implement a dynamics randomization strategy similar to the one used in our previous work [7]. Additionally, we also adopt the same training curriculum as outlined in [7] to further improve the policy's robustness and adaptability.

TABLE III
REWARD TERMS

| | | |
|---|---|---|
| $r^g$ | $\exp(-\lVert v_{xy}^{\text{cmd}} - v_{xy} \rVert_2 /0.15)$ | 1.0 |
| | $\exp(-\lVert \omega_z^{\text{cmd}} - \omega_z \rVert_2 /0.15)$ | 0.5 |
| $r^l$ | $-\lVert \tau \rVert_2$ | $1\times 10^{-4}$ |
| | $-\lVert \ddot{q} \rVert_2$ | $2.5\times 10^{-7}$ |
| | $-\lVert q_{t-1} - q_t \rVert_2$ | 0.1 |
| | $-\lVert h_b^{\text{cmd}} - h_b \rVert_2$ | 1.0 |
| | $-n_{\text{collision}}$ | 0.1 |
| | $\lVert v_z \rVert_2$ | -0.5 |
| | $\lVert \omega_x \rVert_2 + \lVert \omega_y \rVert_2$ | -0.05 |
| | $\lVert g_x \rVert_2 + \lVert g_y \rVert_2$ | -0.5 |
| $r^s$ | (5) | -1.0 |
| $r^c$ | $\sum_{\text{foot}}[1 - C_{\text{foot}}^{\text{cmd}}(\theta^{\text{cmd}}, t)]\exp\{-\lvert f^{\text{foot}} \rvert^2 /\sigma_{cf}\}$ | 1.0 |
| | $\sum_{\text{foot}}[C_{\text{foot}}^{\text{cmd}}(\theta^{\text{cmd}}, t)]\exp\{-\lvert v^{\text{foot}} \rvert^2 /\sigma_{cf}\}$ | 1.0 |

The training process is shown in Fig .6. The model converged after approximately 30,000 training iterations.

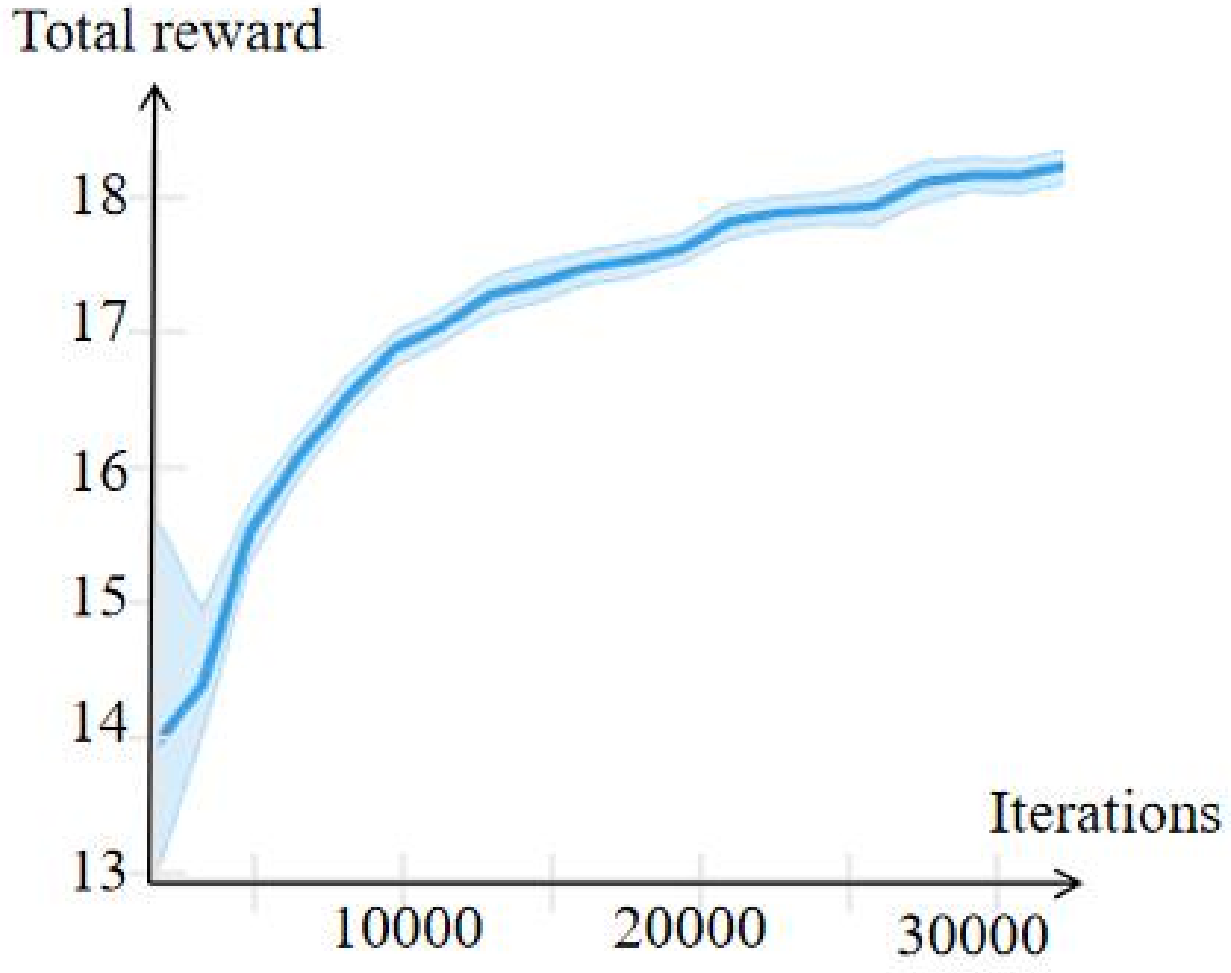

**Fig. 6.** Training convergence curves.

### B. Simulation Results

We conducted extensive experiments in both simulation and real-world environments using the Ask1 and Unitree Go1 robots. These experiments demonstrate that both robots are capable of traversing various rugged terrains, showcasing their versatility and adaptability.

(1) Staircase Environment: In simulated staircase environments, Ask1 showcased its ability to navigate more challenging terrains. The robot successfully traversed stairs with varying heights, demonstrating smooth transitions between steps and the ability to adjust its gait dynamically. This highlighted Ask1's adaptability to complex surfaces, ensuring stable navigation even on irregular terrain. The experimental results are shown in Figure 7.

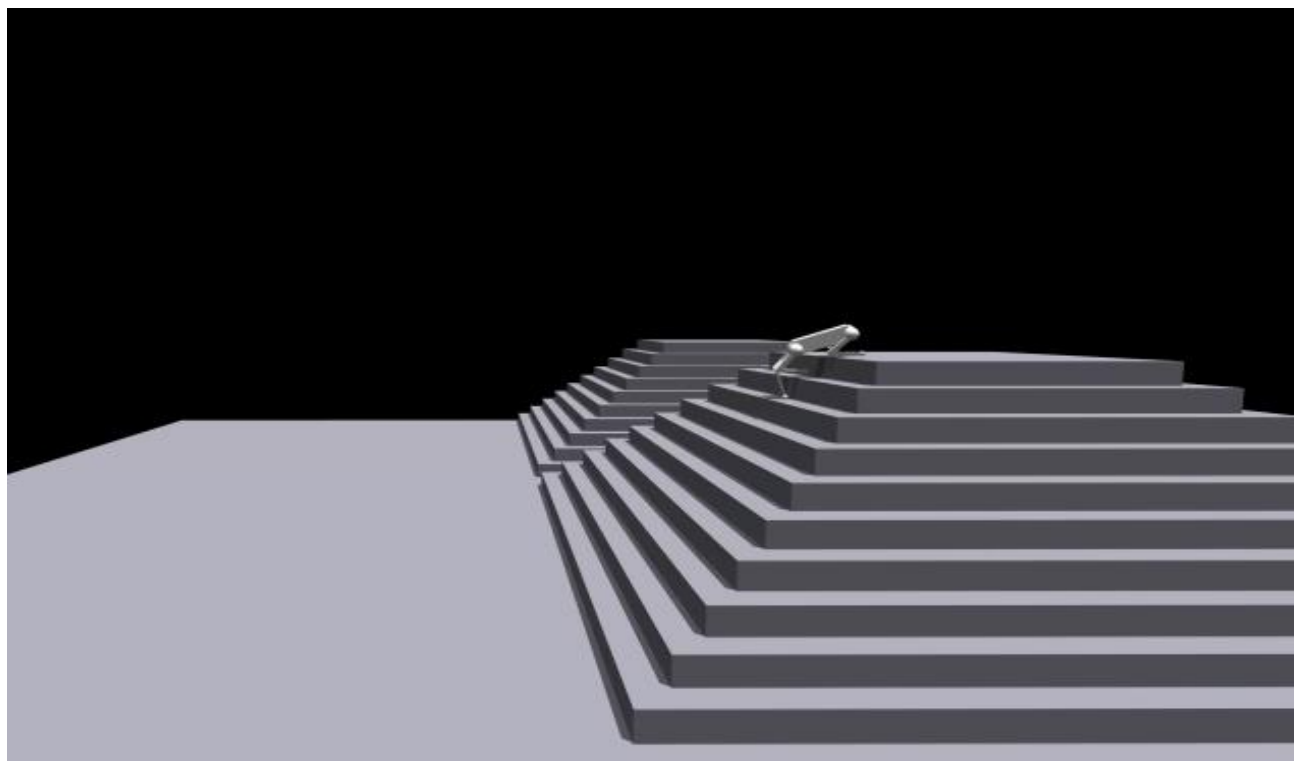

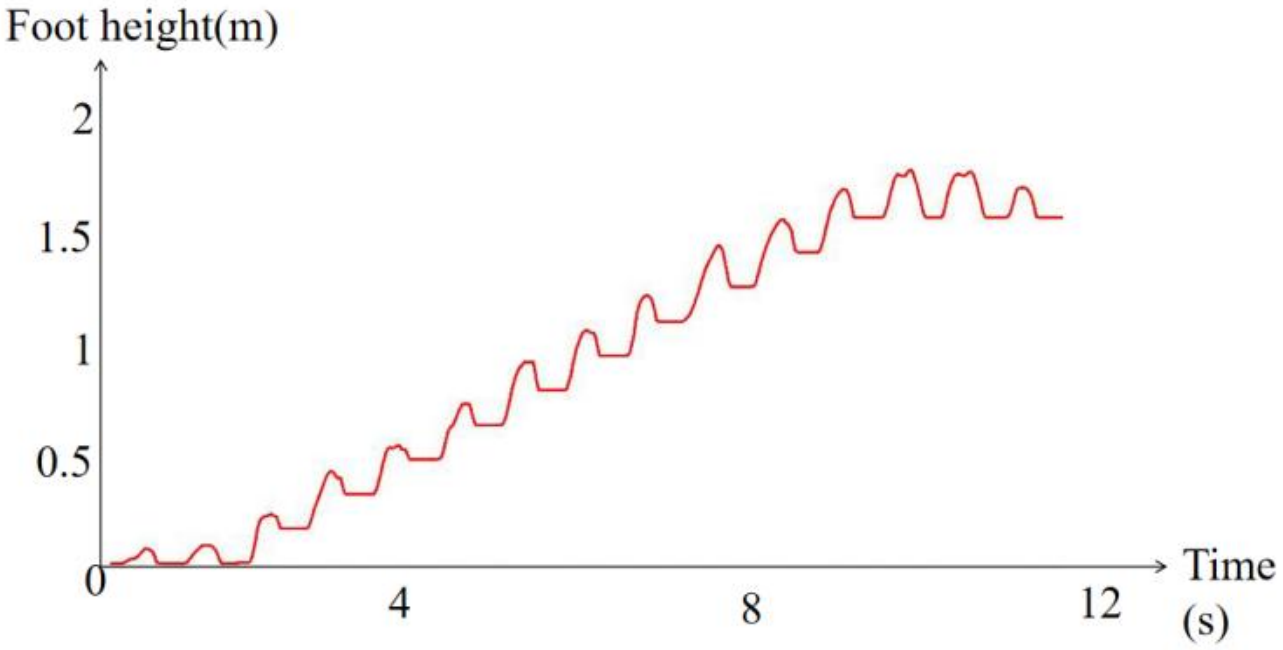

**Fig. 7.** Robot's performance in the simulated stair environment (top) and the variation in foot-end height(bottom). The variation in foot-end height reflects the robot's ability to automatically adjust its leg lift height in response to changes in terrain, ensuring stability and efficient movement across uneven surfaces.

(2) Speed Tracking: We evaluated the speed tracking capabilities of Ask1 on flat terrain, where the robot was tasked with maintaining specific velocities. The results showed that Ask1 could consistently follow speed commands, maintaining stable motion and precise velocity control. This demonstrated its effectiveness in flat, obstacle-free environments. The results are depicted in Fig. 8.

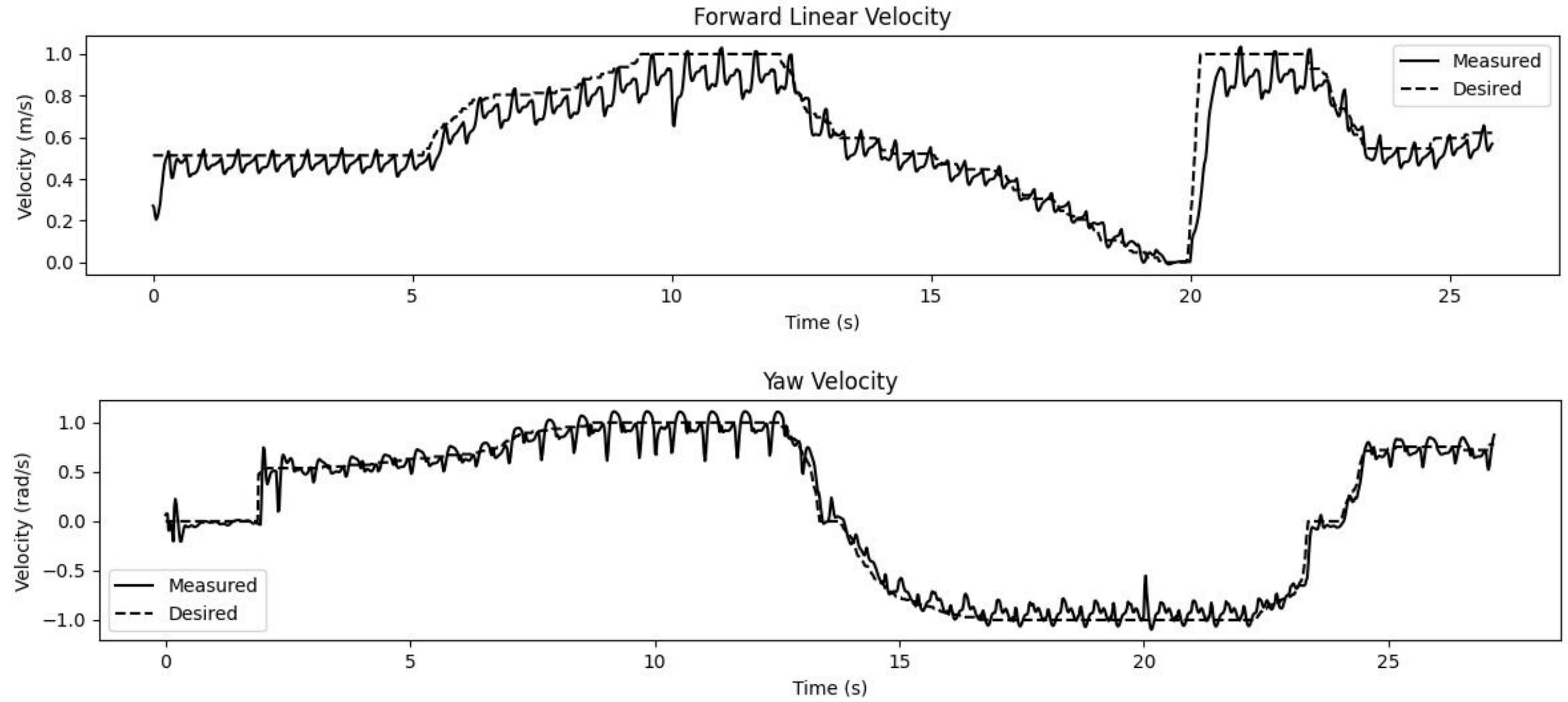

**Fig. 8.** User command tracking: top plot shows linear velocity, bottom plot shows angular velocity.

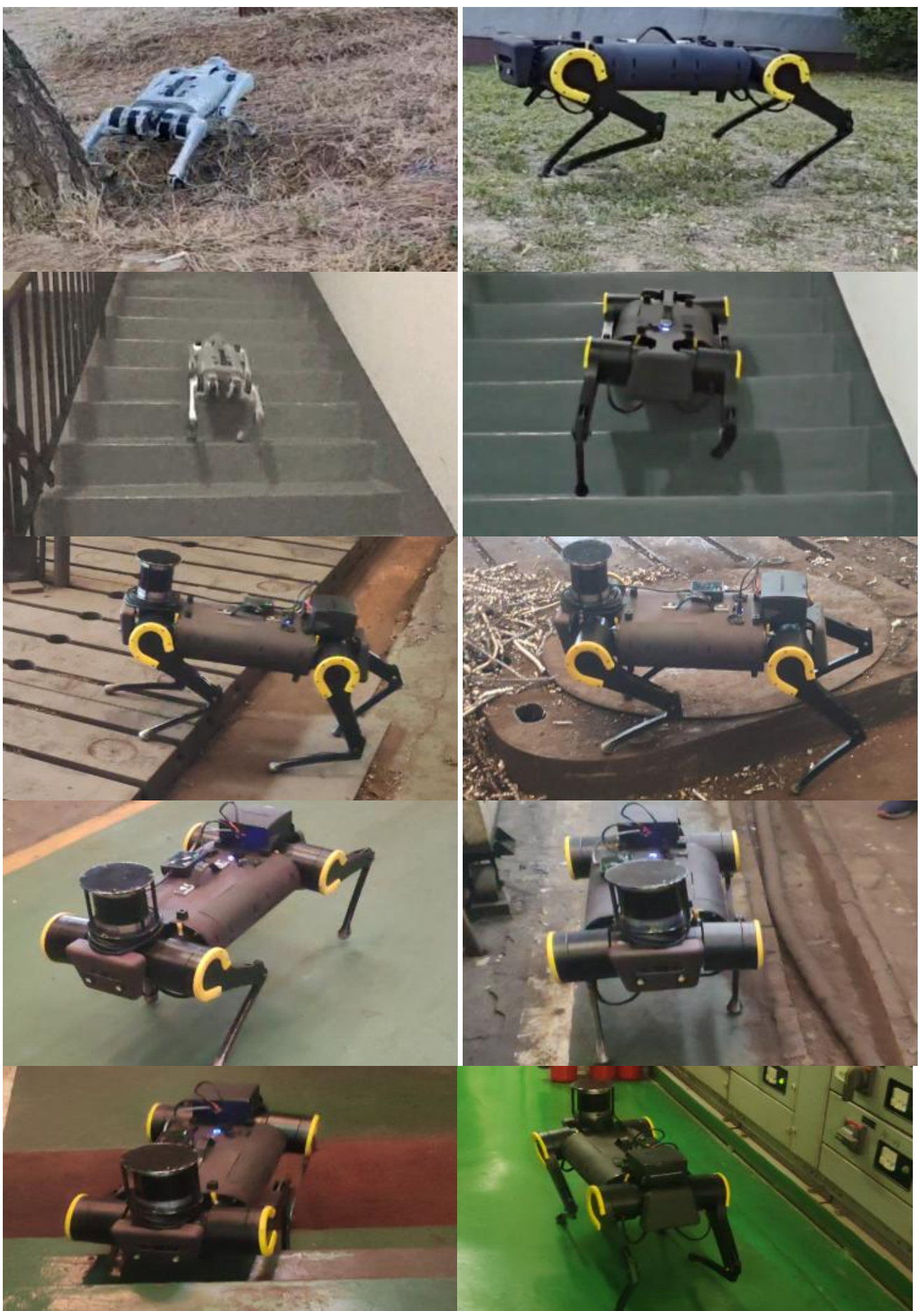

**Fig. 9.** The robots are shown navigating real-world environments: on grass, stairs, and in industrial settings.

*C. Real World Experiments:*

We conducted tests in various outdoor environments. The experimental results are shown in Fig. 9.

(1) Grass Terrain: In real-world testing on grassy surfaces, Ask1 demonstrated exceptional stability and agility, even on uneven ground. The robot maintained a stable gait while adjusting its movement to the natural variations in the grass. This confirmed Ask1's ability to handle softer, unpredictable surfaces.

(2) Staircase Terrain: Real-world tests in staircase environments confirmed that Ask1 could reliably navigate staircases. The robot was able to ascend and descend stairs with ease, adjusting its gait to ensure stability on steep inclines. This shows Ask1's ability to handle vertical challenges, making it suitable for use in environments such as homes, office buildings, or public spaces with stairs.

(3) Industrial Environments: Finally, we tested Ask1 in real industrial environments, such as factory floors and warehouses, where the terrain included ramps, uneven floors, and debris.

The robot performed exceptionally well, navigating across these surfaces with consistent stability. Even when faced with unexpected obstacles, sloped ramps, and uneven floors, Ask1 demonstrated remarkable adaptability, adjusting its gait and balance to maintain smooth and controlled movement. Its ability to seamlessly handle debris, narrow aisles, and varying surface textures highlighted its robustness and versatility in dynamic industrial environments. This performance confirmed that Ask1 is well-suited for real-world industrial applications, proving its reliability in environments that demand both agility and stability.

## V. Conclusion

In this study, we introduced Ask1, a custom quadruped robot, and demonstrated the effectiveness of a reinforcement learning-based control framework for stable and adaptive locomotion. Simulation-trained policies were successfully transferred to the real robot without the need for complex models or additional sensors, allowing Ask1 to perform well in diverse environments, including climbing 18cm stairs. These results showcase the potential of RL for flexible and scalable quadruped robot control.

Future work may explore integrating vision-based inputs and enhancing sensor fusion to further improve adaptability and robustness in more complex environments. This study paves the way for the continued development of autonomous legged robots.

## ACKNOWLEDGE

This work is supported in part by the Liaoning Province Basic and Applied Basic Research Foundation [2023JH2/101600033]; the Guangdong Basic and Applied Basic Research Foundation [2022A1515140156]; the Dalian Science and Technology Innovation Funding (2023JJ12GX017); the State Key Laboratory of Robotics and Systems funding [SKLRS-2023-KF-16]; the National Natural Science Foundation of China under U22B2041.

## References

[1] G. Bledt, M. J. Powell, B. Katz, J. Di Carlo, P. M. Wensing, and S. Kim, “Mit cheetah 3: Design and control of a robust, dynamic quadruped robot,” in 2018 IEEE/RSJ International Conference on Intelligent Robots and Systems (IROS), 2018, pp. 2245–2252.

[2] C. D. Bellicoso, F. Jenelten, C. Gehring, and M. Hutter, “Dy-namic locomotion through online nonlinear motion optimization forquadrupedal robots,” IEEE Robotics and Automation Letters, vol. 3,no. 3, pp. 2261–2268, 2018.

[3] Y. Ding, A. Pandala, C. Li, Y.-H. Shin, and H.-W. Park,“Representation-free model predictive control for dynamic motions in quadrupeds,” IEEE Transactions on Robotics, vol. 37, no. 4, pp.1154–1171, 2021.

[4] B. Ponton, M. Khadiv, A. Meduri, and L. Righetti, “Efficient multicon-tact pattern generation with sequential convex approximations of the centroidal dynamics,” IEEE Transactions on Robotics, vol. 37, no. 5,pp. 1661–1679, 2021.

[5] A. Meduri, P. Shah, J. Viereck, M. Khadiv, I. Havoutis, and L. Righetti,“Biconmp: A nonlinear model predictive control framework for whole body motion planning,” IEEE Transactions on Robotics, 2023.

[6] X. Liu, J. Wu, Y. Xue, C. Qi, G. Xin and F. Gao, "Skill Latent Space Based Multigait Learning for a Legged Robot," in IEEE Transactions on Industrial Electronics.

[7] J. Wu, G. Xin, C. Qi, and Y. Xue, “Learning robust and agile legged locomotion using adversarial motion priors,” IEEE Robotics and Automation Letters, vol. 8, no. 8, pp. 4975–4982, 2023.

[8] G. B. Margolis and P. Agrawal, “Walk these ways: Tuning robotcontrol for generalization with multiplicity of behavior,” Conferenceon Robot Learning, 2022.

[9] X. B. Peng, Z. Ma, P. Abbeel, S. Levine, and A. Kanazawa, “AMP:Adversarial motion priors for stylized physics-based character control,” ACMTrans. Graph., vol. 40, no. 4, pp. 1–20, 2021.

[10] J. Lee, J. Hwangbo, L. Wellhausen, V. Koltun, and M. Hutter,“Learning quadrupedal locomotion over challenging terrain,” Science robotics, vol. 5, no. 47, p. eabc5986, 2020.

[11] A. Kumar, Z. Fu, D. Pathak, and J. Malik, “Rma: Rapid motor adaptation for legged robots,” in Robotics: Science and Systems, 2021.

[12] T. Miki, J. Lee, J. Hwangbo, L. Wellhausen, V. Koltun, and M. Hutter,“Learning robust perceptive locomotion for quadrupedal robots in the wild,” Science Robotics, vol. 7, no. 62, p. eabk2822, 2022.

[13] N. Rudin, D. Hoeller, P. Reist, andM. Hutter, “Learning to walk in minutes using massively parallel deep reinforcement learning,” in Proc. 5th Annu.Conf. Robot Learn., 2021, pp. 91–100.

[14] J. Schulman, F. Wolski, P. Dhariwal, A. Radford, and O. Klimov,“Proximal policy optimization algorithms,” arXiv preprint arXiv:1707.06347, 2017.

[15] J. Hwangbo, J. Lee, A. Dosovitskiy, D. Bellicoso, V. Tsounis, V. Koltun, and M. Hutter, “Learning agile and dynamic motor skills for legged robots,” Science Robotics, vol. 4, no. 26, p. eaau5872, 2019.

[16] G. Margolis, G. Yang, K. Paigwar, T. Chen, and P. Agrawal, “Rapid locomotion via reinforcement learning,” in Robotics: Science and Systems, 2022.

[17] G. Ji, J. Mun, H. Kim, and J. Hwangbo, “Concurrent training of a control policy and a state estimator for dynamic and robust legged locomotion,” IEEE Robotics and Automation Letters, vol. 7, no. 2, pp. 4630–4637, 2022.

[18] L. Pinto, M. Andrychowicz, P. Welinder, W. Zaremba, and P. Abbeel,“Asymmetric actor critic for image-based robot learning,” in Robotics: Science and Systems, 2018.